\pdfoutput=1
\documentclass{article}
\usepackage{spconf,amsmath,graphicx,hyperref,amsfonts,enumerate,multirow,siunitx}

\title{Understanding Textual Capability Degradation in Speech LLMs \\ via Parameter Importance Analysis}
\name{Chao Wang\textsuperscript{*}, Rui-Chen Zheng\textsuperscript{*}\thanks{\textsuperscript{*} Equal Contribution.}, Yang Ai, Zhen-Hua Ling}
\address{National Engineering Research Center of Speech and Language Information Processing,\\
University of Science and Technology of China, Hefei, P. R. China\\
\small \tt \{wangchao2002, zhengruichen\}@mail.ustc.edu.cn,  \{yangai, zhling\}@ustc.edu.cn}
\begin{document}
\ninept
\maketitle
\begin{abstract}
% Recent advancements of speech large language models(speech LLMs) like GPT-4o have brought tremendous improvement of their question-answering capabilities. However, the textual abilities of large language models (LLMs) have declined after the incorporation of the speech modality, resulting in speech LLMs that built on it fail to make the most of the original model's textual competence. To address this, our study employs parameter importance analysis to examine the internal parameters and fine-tuning processes of speech LLMs. The result indicates the internal mechanisms through which the introduction of the speech modality leads to the degradation of textual capability in LLMs. Building on these insights, we propose two methods—layer-wise learning rate scheduling and Low-Rank Adaptation (LoRA)—to improve the fine-tuning process of speech LLMs. These methods successfully enhance the models' question-answering capability. Furthermore, our analysis of different fine-tuning strategies indicates that full fine-tuning tends to cause a shift in the original distribution of textual knowledge within the LLMs, whereas both layer-wise learning rates and LoRA mitigate this effect. By reducing the loss in textual question-answering capability, these approaches consequently lead to improvements in speech question-answering.
The integration of speech into Large Language Models (LLMs) has substantially expanded their capabilities, but often at the cost of weakening their core textual competence. This degradation limits the ability of speech-enabled LLMs to fully exploit their pre-trained text-based knowledge. In this work, we analyze the underlying mechanisms of this issue through a focused study of the widely used encoder–adaptor paradigm. We propose an analytical framework based on parameter importance estimation, which reveals that fine-tuning for speech introduces a textual importance distribution shift: the layer-wise allocation of parameters critical to textual reasoning is disrupted. Building on this insight, we investigate two mitigation strategies: layer-wise learning rate scheduling and Low-Rank Adaptation (LoRA), both aim to preserve the original parameter distribution. Experimental results show that both approaches better maintain textual competence than full fine-tuning, while also improving downstream spoken question answering performance. Furthermore, our analysis offers a principled explanation for the effectiveness of the proposed mitigation strategies, linking their benefits to the structural properties of textual knowledge in LLMs.
\end{abstract}
\begin{keywords}
speech LLM, question answering, parameter importance, textual capability degradation
\end{keywords}
%

% \begin{figure*}[tb]
% %
% \begin{minipage}[t]{0.23\linewidth}
%   \centering
%   \centerline{\includegraphics[width=4cm]{figures/8b/speech_layers.12.mlp.up_proj.png}}
% %  \vspace{1.5cm}
%   \centerline{(a) Speech layers.12.}
%   \centerline{mlp.up\_proj}\medskip
% \end{minipage}
% \hfill
% \begin{minipage}[t]{0.23\linewidth}
%   \centering
%   \centerline{\includegraphics[width=4cm]{figures/8b/text_layers.16.mlp.down_proj.png}}
% %  \vspace{1.5cm}
%   \centerline{(b) Text layers.16.}
%   \centerline{mlp.down\_proj}\medskip
% \end{minipage}
% \hfill
% \begin{minipage}[t]{0.23\linewidth}
%   \centering
%   \centerline{\includegraphics[width=4cm]{figures/8b/speech_layers.25.self_attn.q_proj.png}}
% %  \vspace{1.5cm}
%   \centerline{(c) Speech layers.25.}
%   \centerline{self\_attn.q\_proj}\medskip
% \end{minipage}
% \hfill
% \begin{minipage}[t]{0.23\linewidth}
%   \centering
%   \centerline{\includegraphics[width=4cm]{figures/8b/text_layers.27.self_attn.o_proj.png}}
% %  \vspace{1.5cm}
%   \centerline{(d) Text layers.27.}
%   \centerline{self\_attn.o\_proj}\medskip
% \end{minipage}
% %
% \caption{Parameter importance distribution of different modules(the Top5\% region). The scale from 0 to 1 represent the proportion of parameters within a 3 × 3 vicinity that belong to the ‘Top’ region. ‘Speech’ and ‘Text’ represent the input modality.}
% \label{fig:rank_cluster}
% %
% \end{figure*}

\section{Introduction}
\label{sec:intro}
% With the advancement of large language models(LLMs), how to empower LLMs with speech interaction competence has attracted broad interest among researchers. For example, GPT-4o enables real-time, intelligent, and natural speech interaction between users and LLMs.
Large language models (LLMs) have achieved remarkable success across a wide range of natural language processing tasks \cite{achiam2023gpt, team2023gemini, dubey2024llama, yang2025qwen3}. Building on this foundation, enabling LLMs to operate across multiple modalities, particularly speech, has emerged as a key research frontier. The recent release of GPT-4o \cite{hurst2024gpt} highlights the transformative potential of this paradigm, demonstrating highly effective real-time and natural speech interaction. The dominant approach for developing such systems is not to train them from scratch, but rather to adapt powerful, pre-trained text-based LLMs to the speech modality, thereby enabling both speech understanding and generation \cite{arora2025landscape}.

Early attempts at speech-enabled LLMs often relied on cascaded ASR–LLM–TTS pipelines \cite{zhang-etal-2023-speechgpt, AudioGPT}. While functional, these pipelines introduce non-negligible latency and suffer from error accumulation across components. To overcome these limitations, recent work has shifted toward end-to-end speech LLMs that directly integrate speech abilities. Existing methods generally fall into two paradigms. The first couples a speech encoder with the LLM via an adaptor, aligning speech representations with LLM's input space \cite{llamaomni, freeze-omni, minmo}. The second directly incorporates discrete speech tokens into the LLM by expanding its vocabulary \cite{defossez2024moshi, mini-omni, mini-omni2, glm4}.

To equip LLMs with speech capabilities, both paradigms typically rely on multi-stage fine-tuning on speech-related tasks \cite{peng2025voicetextblender}. However, the substantial mismatch between speech and text data distributions often leads to catastrophic forgetting, a phenomenon where the LLM loses previously acquired knowledge \cite{lu25c_interspeech, hsiao25_interspeech}. While the LLM gains new speech-related skills, its foundational text-based reasoning and instruction-following abilities significantly degrade\cite{fang2025s2sbench}. This trade-off poses a critical bottleneck, as preserving the strong textual intelligence of the base LLM is essential for high-quality spoken interaction. Although this degradation is widely observed, its internal mechanisms remain poorly understood. What exactly is disrupted in the model during speech fine-tuning, and why does this lead to diminished textual competence?

% To investigate the reasons behind the performance gap in question-answering capabilities between speech LLMs and original LLMs, we employed parameter importance method to analyze the internal parameters and fine-tuning processes of speech LLMs. Experimental results indicates the internal mechanisms through which the introduction of the speech modality leads to a degradation in the textual capabilities of large language models.

% The parameter importance analysis of speech LLMs shows that the parameter importance distribution displays an concentration in both the rows and columns of the parameter matrices, which we called rank clustering. Deactivating (zeroing out) only the top 3\% most important parameters causes a sharp increase in the model’s PPL and nearly eliminates its linguistic competence, while deactivating either random 3\% or the bottom 3\% of parameters has almost no impact on model performance. Moreover, by examining the layer-wise distribution of parameter importance, we found that full fine-tuning shifts the text-related parameter importance of the original model, resulting in reduced textual capability.

To address these questions, this paper conducts a focused case study of speech LLMs built with the encoder–adaptor paradigm \cite{llamaomni}. We select this architecture not only for its widespread adoption but also for its methodological advantage: the adaptor projects speech representations into the textual embedding space while leaving the base LLM unchanged. This structural separation enables a controlled analysis of how fine-tuning affects the model’s internal parameters, without the confounding input modifications introduced by vocabulary expansion. Building on this setup, we develop an analytical framework based on parameter importance estimation \cite{ImportanceEstimation,unveiling-linguistic} to quantify the sensitivity of individual parameters and examine their layer-wise distribution. 
Through this analysis, we identify a key mechanism underlying textual degradation: fine-tuning with speech induces a shift in the original distribution of parameter importance, thereby disrupting the model’s textual competence.

% Based on these findings, we propose two methods to mitigate the damage to textual ability during fine-tuning:
% (1)Layer-wise learning rate scheduling based on the textual parameter importance of the original model.Layers with higher textual importance are assigned lower learning rates to reduce distribution shift.
% (2)Leveraging the row/column-wise concentration of parameter importance, we hypothesize that optimizing parameter matrices in a structured manner (e.g., by rows/columns) better aligns with the model’s adaptation to speech-modal tasks. Hence, we apply LoRA fine-tuning.
Building on this analysis, we adopt two strategies to mitigate textual degradation: layer-wise learning rate scheduling \cite{layer-lr}, which adjusts updates across layers to better preserve textually important parameters, and Low-Rank Adaptation (LoRA) \cite{lora}, which constrains updates within a low-rank subspace to minimize disruption to pre-trained parameters.
% Using answer accuracy on dual-modality (text and speech) question-answering datasets as the metric, we validated the effectiveness of both methods. Furthermore, analysis across different fine-tuning approaches indicates that full fine-tuning distorts the original textual knowledge distribution of the LLM, whereas both layer-wise learning rate scheduling and LoRA fine-tuning help mitigate this effect, reduce the loss in textual QA performance, and thereby improve speech-based QA capability.
Extensive experiments on dual-modality question answering benchmarks \cite{llamaQA, webQA} demonstrate that both methods outperform full fine-tuning in maintaining textual competence while simultaneously improving speech comprehension. Furthermore, our analysis explains why these strategies are effective: layer-wise scheduling reduces distribution shifts, while LoRA aligns with the inherent low-rank structure of textual knowledge.

\begin{figure}[tb]
    \centering
    \includegraphics[width=0.9\linewidth]{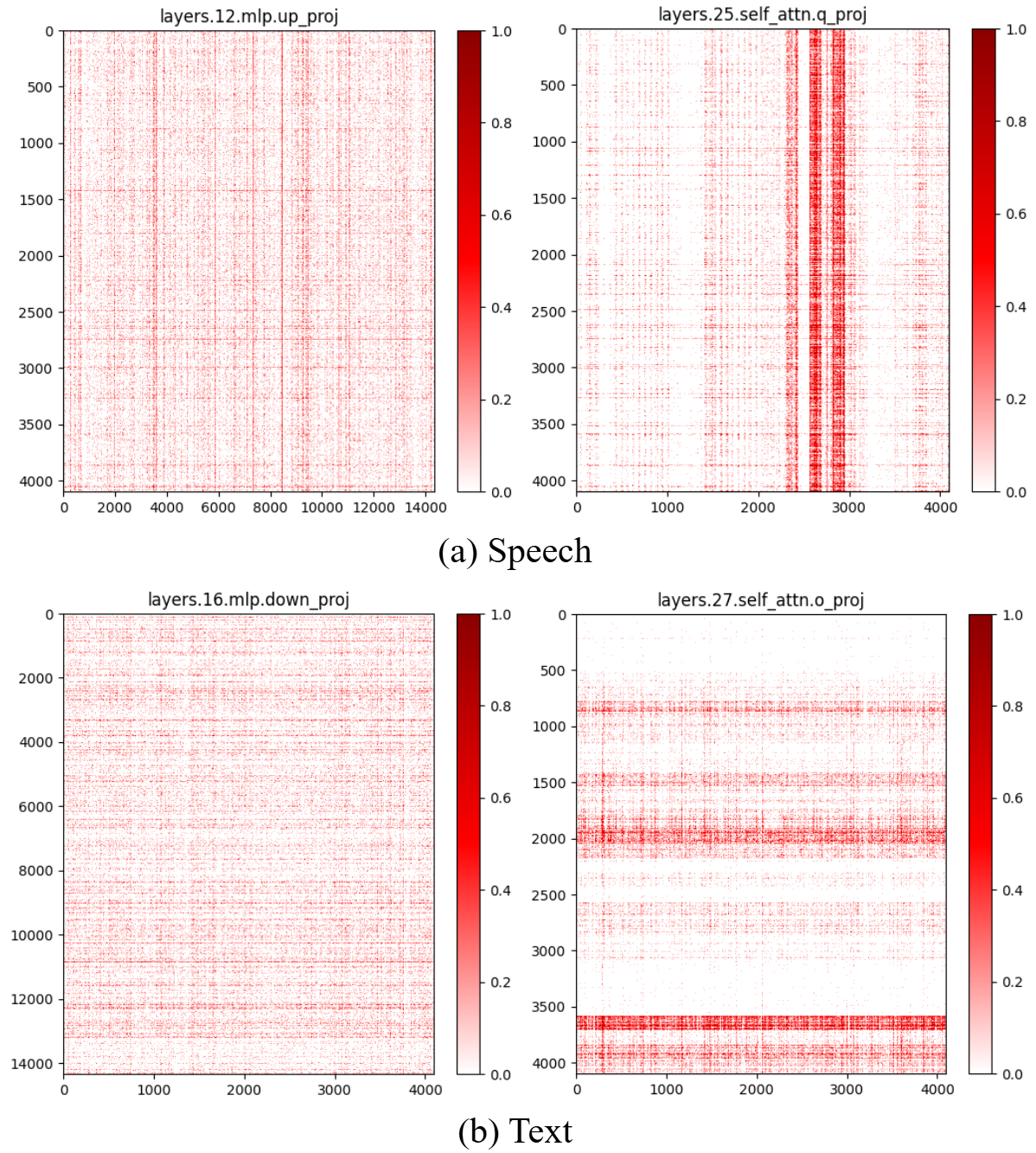}
    \caption{Parameter importance distribution of different modules (Top5\% region) with different input modality. The scale from 0 to 1 (after normalization) represent the proportion of parameters within a 3 × 3 vicinity that belong to the ‘Top’ region.}
\label{fig:rank_cluster}
\end{figure}

% \section{Analysis of Speech LLMs using Parameter Importance}
\section{A Parameter-Level Analysis of Textual Capability Degradation in Speech LLMs}
\label{sec:analysis}

\subsection{Analytical Framework}
\label{ssec:param_imp}
Our analysis is grounded in the LLaMA-Omni architecture \cite{llamaomni}, a representative example of the encoder-adaptor paradigm. We focus on this paradigm both for its prevalence and for its suitability as a controlled setting: the speech front-end (encoder and adaptor) projects acoustic features into the LLM’s representation space, leaving the vocabulary and embedding layers unchanged. This separation allows us to attribute changes in textual competence directly to the fine-tuning process, avoiding the confounding effects present in vocabulary expansion. In practice, the encoder–adaptor design has been widely adopted in state-of-the-art systems \cite{minmo, xu2025qwen2}, ensuring that our findings are broadly relevant.

For this study, implemented the architecture with two different base models: LLaMA-3.2-1B and LLaMA-3.1-8B. Our analysis concentrates on the first stage of the LLaMA-Omni training pipeline, which fine-tunes the LLM for speech understanding. This stage is most critical for our research question, as the subsequent speech generation stage freezes the LLM parameters and thus does not affect its intrinsic textual knowledge.
To probe the internal changes that occur during this stage, we employ parameter importance estimation \cite{ImportanceEstimation, unveiling-linguistic}. This method measures the contribution of each parameter to the model’s linguistic performance by quantifying the sensitivity of the loss to that parameter. Specifically, the importance $I_i(\theta)$ of a parameter $\theta_i$ is formally defined as the absolute change in model loss when $\theta_i$ is nullified (i.e., set to zero):
\begin{equation}
I_i(\theta) = |L(D, \theta) - L(D, \theta|\theta_i = 0)|,
\end{equation}
where $L(D, \theta)$ denotes the loss over a given dataset $D$. As calculating this value for every parameter across a large model is computationally infeasible, we approximate it using a first-order Taylor expansion, yielding the widely used gradient-based estimate \cite{ImportanceEstimation}:
\begin{equation}
I_i(\theta) \approx |\frac{\partial L}{\partial \theta_i}\theta_i|.
\end{equation}
This  approximation provides a reliable estimate of parameter importance and serves as the foundation for our subsequent analysis.

\subsection{Validation of the Parameter Importance Estimation}
\label{ssec:Exp_Setup}
To validate the reliability of parameter importance estimation, we conducted a deactivation experiment using both the LLaMA-Omni-1B and LLaMA-Omni-8B model. Our hypothesis was that parameters with high importance scores would be disproportionately critical to the model’s linguistic functions. 
The models were trained on first-turn dialogues from the VoiceAssistant-400K \footnote{\url{https://huggingface.co/datasets/gpt-omni/VoiceAssistant-400K}} and Spoken-Alpaca-GPT4 \footnote{\url{https://huggingface.co/datasets/GSQA/spoken-alpaca-gpt4}} datasets, where each spoken query was paired with its textual transcription to enable controlled comparison across modalities. For evaluation, we constructed a test set of 500 randomly sampled examples that were disjoint from the training data.
As our focus is on the impact of speech integration on textual ability, the model was evaluated by generating text responses from both speech and text inputs.
To assess the effect of parameter removal, we compared text generation performance under three conditions: the top 3\% with the highest importance scores, the bottom 3\% with the lowest scores, and a randomly selected 3\%. Parameter importance scores were computed using a 1/30 subset of the training data, and model perplexity (PPL) on the test set was used as the evaluation metric.

The results, summarized in Table \ref{tab:deact}, strongly support our hypothesis. Deactivating the top 3\% of parameters caused a dramatic increase in PPL and a severe loss of linguistic competence. In contrast, nullifying the bottom 3\% had negligible effect, and removing a random 3\% resulted in only minor degradation. These findings demonstrates that the observed performance degradation is attributed to the removal of high-importance parameters rather than the deactivation process itself. We therefore conclude that the parameter importance metric is a reliable tool for identifying functionally critical parameters, providing a solid foundation for the subsequent analysis.

\begin{table}[t]
    \centering
    \caption{Model perplexity (PPL) after deactivating 3\% parameters based on parameter importance scores. }
    \begin{tabular}{c @{\hspace{0.6em}} c c c c c}
        \hline
        \multirow{2}{*}{Size} & \multirow{2}{*}{Input} & \multirow{2}{*}{Base} & \multicolumn{3}{c}{3\% Removal} \\ \cline{4-6}
        &  &  & Top & Bottom & Random \\ \hline 
        \multirow{2}{*}{1B} & Speech & 2.08 & \num{1.14e5} & 2.18 & 3.85 \\
        & Text & 3.65 & \num{2.68e5} & 3.81 & 6.39 \\ \hline 
        \multirow{2}{*}{8B} & Speech & 1.75 & \num{2.72e5} & 1.76 & 3.54 \\ 
        & Text & 3.12 & \num{2.60e5} & 3.19 & 5.47 \\ \hline
    \end{tabular}
    \label{tab:deact}
\end{table}

% \subsection{Rank Clustering}
% \label{ssec:dim_dep}
% To explore the distribution of linguistic competence within the model modules, we plot the parameter importance heat map of different modules. We use Speech-Llama-8B as the example. As shown in figure \ref{fig:rank_cluster}, whether in the self-attention matrix or the feed-forward matrix, the parameter importance distribution displays an obvious concentration in both the rows and columns of the matrices, which we called rank clustering. This distribution feature may indicate that the linguistic competence of the model depends on specific rows and columns.

\subsection{Core Findings}
With the parameter importance metric validated, we applied it to analyze the distribution of parameters in speech LLMs. This analysis yields two key findings. First, we observe a consistent structural pattern in the organization of important parameters. Second, we identify a shift in this distribution caused by fine-tuning with speech inputs, which we regard as the primary mechanism of textual degradation.

\subsubsection{Structural Pattern: Rank Clustering}
\label{ssec:dim_dep}
When visualized as heatmaps of Transformer weight matrices, parameter importance exhibits a structural pattern: important parameters cluster along certain rows and columns, a phenomenon we term \textit{rank clustering}. As shown in Figure \ref{fig:rank_cluster}, important parameters are not randomly scattered but concentrated along specific rows and columns. This distribution is consistent with the importance patterns observed in text LLMs \cite{unveiling-linguistic}. The non-uniform allocation suggests that the model’s linguistic competence is encoded in a low-rank structure, rather than being evenly spread across parameters.

\subsubsection{Textual Importance Distribution Shift}
\label{sssec:text_dis_shift}
% Previous studies have indicated that full fine-tuning of speech LLMs can impair the model's original textual capabilities. 
% To investigate the specific mechanism through which full fine-tuning affects the model's textual ability, we compare the layer-wise distribution of textual parameter importance before and after fine-tuning. The results indicate that a distribution shift occurs during the fine-tuning process.

\begin{figure}[t]
    \centering
    \includegraphics[width=0.95\linewidth]{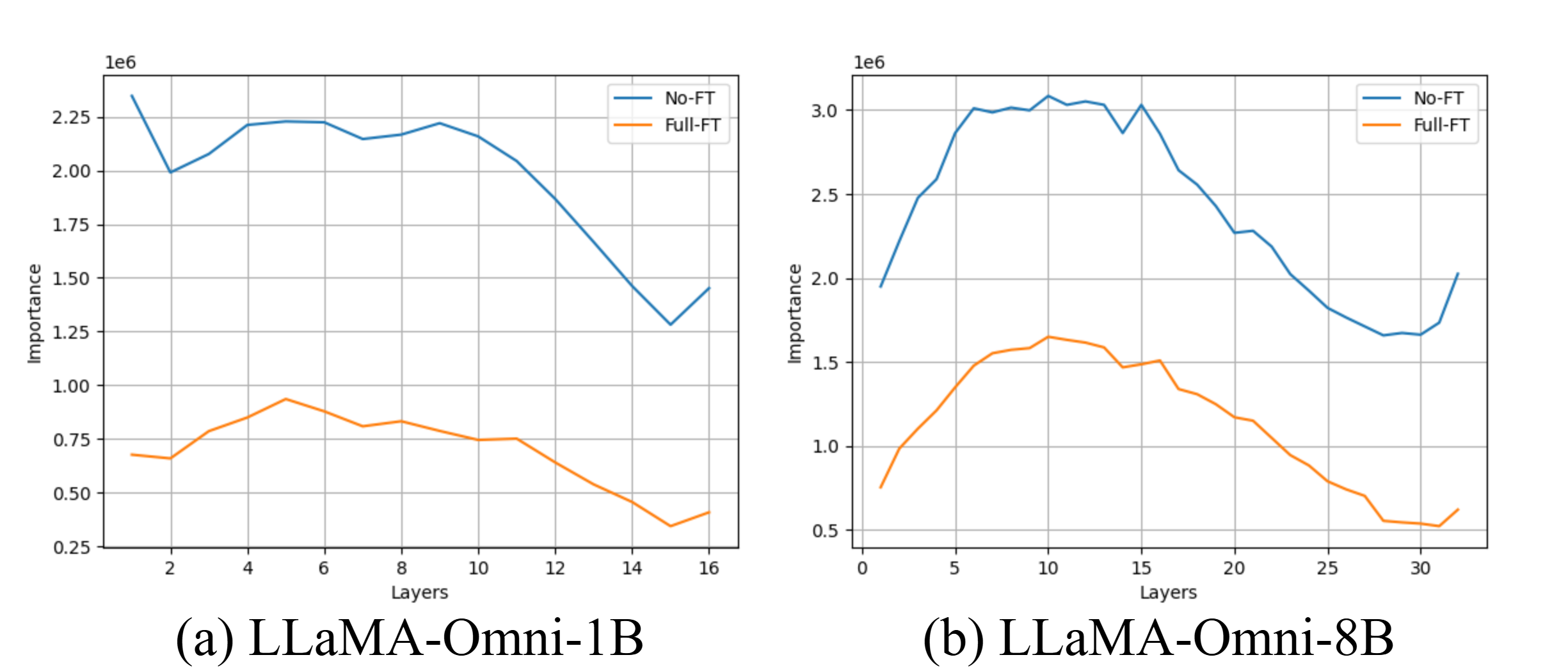}
    \caption{Layer-wise distribution of textual parameter importance before and after full fine-tuning.}
    \label{fig:imp_dist_shift}
\end{figure}

% Specifically, we aggregated the absolute values of textual parameter importance for the MLP parameters within each layer, using the sum as the total textual parameter importance per layer. We then plotted the layer-wise distribution of these values. As shown in Figure \ref{fig:imp_dist_shift}, in the Speech-Llama-1B, the textual parameter importance decreases consistently starting from the fifth layer, whereas in the pre-fine-tuning model, the decrease only begins from the ninth layer. In other words, the textual parameter importance in the later layers declines more rapidly in the fine-tuned model. The similar situation also exists in Speech-Llama-8B. This distribution shift may be one of the reasons for the degradation in textual performance after fine-tuning.
While rank clustering captures the static structure of parameter importance, our central question concerns how this structure changes during speech adaptation. To this end, we compared the layer-wise distribution of textual parameter importance before and after fine-tuning  with speech inputs. The results, illustrated in Figure \ref{fig:imp_dist_shift}, reveal a consistent distribution shift. In the 1B base model, textual importance is concentrated in the later-middle layers, which are typically associated with higher-level semantic processing. In the 8B model, the peak importance lies in the earlier-middle layers and gradually decreases in deeper layers. After fine-tuning on speech input, this distribution is disrupted. In the 1B model, the shift is particularly severe: the peak of importance moves to much earlier layers, followed by a sharp decline. In the 8B model, the shape of the distribution is better preserved, but the overall magnitude of importance is suppressed across all layers. Despite these scale-dependent differences, the outcome is consistent: the relative importance of middle and last few layers is significantly diminished in both models.
We posit that this layer-wise distribution shift is the primary internal mechanism responsible for the degradation of textual performance when LLMs are fine-tuned to incorporate speech.

% \begin{figure}[t]
% %
% \begin{minipage}[b]{0.48\linewidth}
%   \centering
%   \centerline{\includegraphics[width=4.0cm]{figures/tensor_imp_layers_1b.png}}
% %  \vspace{1.5cm}
%   \centerline{(a) Speech-Llama-1B}\medskip
% \end{minipage}
% \begin{minipage}[b]{0.48\linewidth}
%   \centering
%   \centerline{\includegraphics[width=4.0cm]{figures/tensor_imp_layers_8b.png}}
% %  \vspace{1.5cm}
%   \centerline{(b) Speech-Llama-8B}\medskip
% \end{minipage}
% %
% \caption{Layer-wise distribution of textual parameter importance before and after full fine-tuning.}
% \label{fig:imp_dist_shift}
% %
% \end{figure}

\section{Mitigation of Textual Capability Degradation}
\label{sec:improvement}

\subsection{Mitigation Strategies}
\label{ssec:improve}
% To mitigate the aforementioned issues, we validated the correctness of our conclusions through two intuitive methods: layer-wise learning rate scheduling\cite{layer-lr} and Low Rank Adaptation(LoRA)\cite{lora}.
Based on our analysis in Section \ref{sec:analysis}, which identifies the textual importance distribution shift as the primary cause of degradation, we propose and evaluate two mitigation strategies designed to preserve the model's original knowledge structure during speech adaptation.

\subsubsection{Layer-wise Learning Rate Scheduling} 
\label{sssec:Layer-LR}
% The first method is layer-wise learning rate scheduling(Layer-LR): We hypothesized that modifying the model parameters with smaller update magnitudes would better preserve the original capabilities of the model. Therefore, we set the learning rate for each layer based on its parameter importance, assigning lower learning rates to layers with higher importance. Specifically, the learning rate for each layer is multiplied by a specific coefficient, and the coefficient for layer $i$ can be expressed by the following formula:
Our first approach directly targets the distribution shift observed in Section \ref{sssec:text_dis_shift}. We hypothesize that preserving the importance of critical layers can be achieved by reducing the magnitude of their updates during fine-tuning. Therefore, we implement a layer-wise learning rate scheduling strategy \cite{layer-lr} where layers with higher textual parameter importance (as measured on the pre-trained model) are assigned lower learning rates. The learning rate for each layer is scaled by a coefficient calculated as follows:
\begin{equation}
\begin{aligned}
lr(i) = 1 - \lambda * \frac{I_{\text{layer}}(i) - min(I_{\text{layer}}(i))}{max(I_{\text{layer}}(i)) - min(I_{\text{layer}}(i))},
\end{aligned}
\end{equation}
where $I_{\text{layer}}(i)$ is the sum of the absolute importance values of all parameters in layer $i$, and $\lambda$ is a scaling factor empirically set to 0.4. 
% All other training settings remain identical to the full fine-tuning baseline.

\subsubsection{Low Rank Adaptation}
\label{sssec:LoRA}
% The second method involves the use of LoRA fine-tuning. On the one hand, LoRA fine-tuning has been widely adopted and empirically validated to better preserve the model's original capabilities. On the other hand, as discussed in section \ref{ssec:dim_dep}, the parameter importance distribution in speech LLMs exhibits the phenomenon of rank clustering. We thus hypothesize that a rank-based optimization approach is more aligned with the model’s adaptation to speech inputs—a requirement just met by LoRA fine-tuning.

% For the layer-wise learning rate method, all training settings remain consistent with full fine-tuning, with the exception of the additional layer-wise learning rate coefficients.
% For LoRA fine-tuning, the targets are all the MLP and self-attention modules. For the 1B and 8B models, rank is set to 8 and 16, respectively, and the LoRA scaling factor is set to twice the rank value. The learning rate is set to 2e-4, while all other settings matched those of full fine-tuning.
% The finetuned models are named using the convention:  Speech-Llama-{size}-{method}.

% Our second approach is motivated by the "rank clustering" phenomenon identified in Section \ref{ssec:dim_dep}. The observation that important parameters are concentrated in a low-rank structure suggests that a rank-structured optimization method would be particularly well-suited for speech adaptation. LoRA \cite{lora}, which constrains parameter updates to low-rank decomposition matrices, aligns perfectly with this hypothesis. 
The rank clustering phenomenon identified in Section \ref{ssec:dim_dep} offers a principled explanation for the effectiveness of LoRA \cite{lora} in the context of speech LLMs. Our finding that important parameters are concentrated in a low-rank structure suggests that an ideal adaptation method should respect this inherent structure to avoid catastrophic forgetting. LoRA, which constrains parameter updates to low-rank decomposition matrices, aligns perfectly with this principle. Therefore, we leverage LoRA not merely as a parameter-efficient fine-tuning technique, but as a method whose core mechanism is uniquely suited to preserving the essential textual knowledge structure we have identified. For our experiments, we apply LoRA to all MLP and self-attention modules, with a rank $r=8$ for the 1B model and $r=16$ for the 8B model, and the scaling factor $\alpha$ set to $2r$.

\begin{table}[t]
    \centering
    \caption{Results on spoken QA benchmarks. * indicates the original results reported in \cite{llamaomni}. \textbf{Bold} numbers indicate the best performance, and \underline{underlined} numbers indicate the suboptimal.}
    \resizebox{\linewidth}{!}{%
    \begin{tabular}{l c c c c c c}
        \hline
        \multirow{2}{*}{Model} & \multirow{2}{*}{Size} & \multirow{2}{*}{Methods} & \multicolumn{2}{c}{Llama Q} & \multicolumn{2}{c}{Web Q} \\
         & & & T2T & S2T & T2T & S2T \\ \hline 
         Moshi & 7B & \multirow{3}{*}{Full-FT} & - & 62.3 & - & 26.6 \\
         GLM-4-Voice & 9B & & - & 64.7 & - & 32.2 \\
         LLaMA-Omni* & 8B & & - & 67.7 & - & 33.4 \\ \hline
         \multirow{4}{*}{LLaMA-Omni} & \multirow{4}{*}{1B} & No-FT & \textbf{74.0}& - & \textbf{44.5}& - \\
         & & Full-FT & 73.3 & 66.7 & 42.1 & 29.1 \\ 
         & & Layer-LR & \underline{73.7}& \underline{68.3}& \underline{43.8}& \underline{30.2}\\
         & & LoRA & \underline{73.7}& \textbf{70.3}& 42.9 & \textbf{33.5}\\ \hline
         \multirow{4}{*}{LLaMA-Omni}& \multirow{4}{*}{8B} & No-FT & \textbf{84.7}& - & \textbf{58.7}& - \\
         & & Full-FT & 80.0 & 72.0 & 55.7 & 38.7 \\ 
         & & Layer-LR & \underline{81.3}& \underline{73.3}& \underline{57.6}& \underline{39.6}\\
         & & LoRA & 81.0 & \textbf{75.0}& 56.7 & \textbf{42.9}\\ \hline
    \end{tabular}
    }
    \label{tab:improve_res}
\end{table}

\subsection{Results}
\label{ssec:results}
% To observe the mitigating effect of the proposed method on the text capability degradation issue in full fine-tuning, we evaluated the models using their answer accuracy on question-answering datasets. Specifically, the datasets used are Llama Questions\cite{llamaQA} and Web Questions\cite{webQA}, along with their spoken versions. The answer accuracy for text input is used to represent the model's textual capability, while the answer accuracy for speech input with identical content is used to represent the model's speech capability. 
% To evaluate the proposed two mitigation strategies, we evaluate our proposed mitigation strategies on two spoken question answering benchmarks: Llama Questions\footnote{\url{https://huggingface.co/datasets/fixie-ai/llama-questions}} \cite{llamaQA} and Web Questions\footnote{\url{https://huggingface.co/datasets/chiyuanhsiao/spoken-web-questions}} \cite{webQA}, using a fully fine-tuned (Full-FT) model as the primary baseline. Performance is measured on two tasks designed to have identical content: a Text-to-Text (T2T) task to assess the preservation of textual capabilities, and a Speech-to-Text (S2T) task to assess the acquired speech comprehension.
All models were trained on the datasets introduced in Section \ref{ssec:Exp_Setup}. Except for the parameter configurations specified in Section \ref{ssec:improve}, all other settings followed the Stage 1 setup of the original LLaMA-Omni paper \cite{llamaomni}. This ensures that any performance differences can be attributed to the proposed strategies rather than unrelated implementation details. Evaluation was conducted on two spoken question answering (QA) benchmarks: LLaMA Questions\footnote{\url{https://huggingface.co/datasets/fixie-ai/llama-questions}} \cite{llamaQA}  and Web Questions\footnote{\url{https://huggingface.co/datasets/chiyuanhsiao/spoken-web-questions}} \cite{webQA}. We evaluated the models on two tasks with identical question–answering content. The \textit{Text-to-Text (T2T)} task corresponds to standard text-based QA, which measures the preservation of textual competence. The \textit{Speech-to-Text (S2T)} task corresponds to spoken QA, where the model receives speech as input and generates textual answers, thereby assessing speech comprehension.  The fully fine-tuned model (\textit{Full-FT}) and the text-only LLM without speech adaptation (\textit{No-FT}) were used as the primary baselines. In addition, we include S2T results from several open-source speech LLMs, such as Moshi \cite{defossez2024moshi} and GLM-4-Voice \cite{glm4}, as well as the results reported by the authors of \cite{llamaomni}, with all numbers directly taken from \cite{llama-omni2}. These comparisons provide a broader context for assessing the performance of our models.

% As shown in table \ref{tab:improve_res}, compared to previous advanced open-source models, our smaller model (1B) already achieves comparable performance, while the larger model (8B) shows a significant improvement in accuracy on speech-based question answering. When comparing the text question answering accuracy before and after fine-tuning, we observed a decline in accuracy regardless of the fine-tuning strategy used, indicating that the fine-tuning process impairs the model's original textual capabilities. In contrast, when comparing models fine-tuned with layered learning rates and LoRA against those using full fine-tuning, both methods exhibited improvements in accuracy on both text and speech question answering tasks. This suggests that the two proposed methods are more effective at preserving the model's inherent textual intelligence, thereby also enhancing its speech comprehension capabilities.
The results, presented in Table \ref{tab:improve_res}, demonstrate the effectiveness of both the Layer-LR and LoRA strategies. Compared with the Full-FT baseline, both strategies substantially mitigate textual degradation while simultaneously improving speech performance. For example, on the Web Questions benchmark, the 8B model trained with full fine-tuning shows a decrease in T2T accuracy from 58.7\% to 55.7. In contrast, the models trained with layer-wise learning rate scheduling and LoRA both achieve higher accuracies of 57.6\% and 56.7\%, respectively, confirming their superior ability to preserve the model's original textual competence. 
This preservation of textual knowledge translates directly to improved spoken QA performance. On the Web-Questions benchmark, the 8B model trained with LoRA achieves the highest accuracy of 42.9\%. These results strongly suggest that retaining the foundational textual abilities of the base LLM is essential for enhancing its speech comprehension capabilities. 

We also observe a noteworthy trade-off: compared to the model fine-tuned with LoRA, the one using the layer-wise learning rate scheduling exhibits a greater improvement in T2T accuracy, while the LoRA-based model achieves higher accuracy on S2T. This indicates that layer-wise learning rate scheduling is more effective at preserving the base model’s textual knowledge, but its conservative updates limit adaptation to speech inputs, leading to smaller improvements in spoken QA. In contrast, LoRA facilitates more efficient transfer of knowledge from text to speech, yielding superior spoken QA performance at the cost of weaker textual preservation. Overall, these findings highlight that different strategies strike different balances between textual competence and speech adaptation.
% Layer-wise learning rate scheduling is better suited for emphasizing textual robustness, whereas Low-Rank Adaptation is more effective for boosting speech capability. Future work may explore hybrid or adaptive strategies to balance these complementary strengths.
% We hypothesize that this phenomenon may be related to rank clustering mentioned in Section \ref{ssec:dim_dep}, which we analyze next.

% It is noteworthy that, compared to the model fine-tuned with LoRA, the one using the layer-wise learning rate method exhibits a greater improvement in text question answering accuracy, while the LoRA-based model achieves higher accuracy on speech question answering. We hypothesize that this phenomenon may be related to rank clustering mentioned in Section \ref{ssec:dim_dep}, with specific details discussed in Section \ref{ssec:analysis}.

\begin{figure}[tb]
    \centering
    \includegraphics[width=0.95\linewidth]{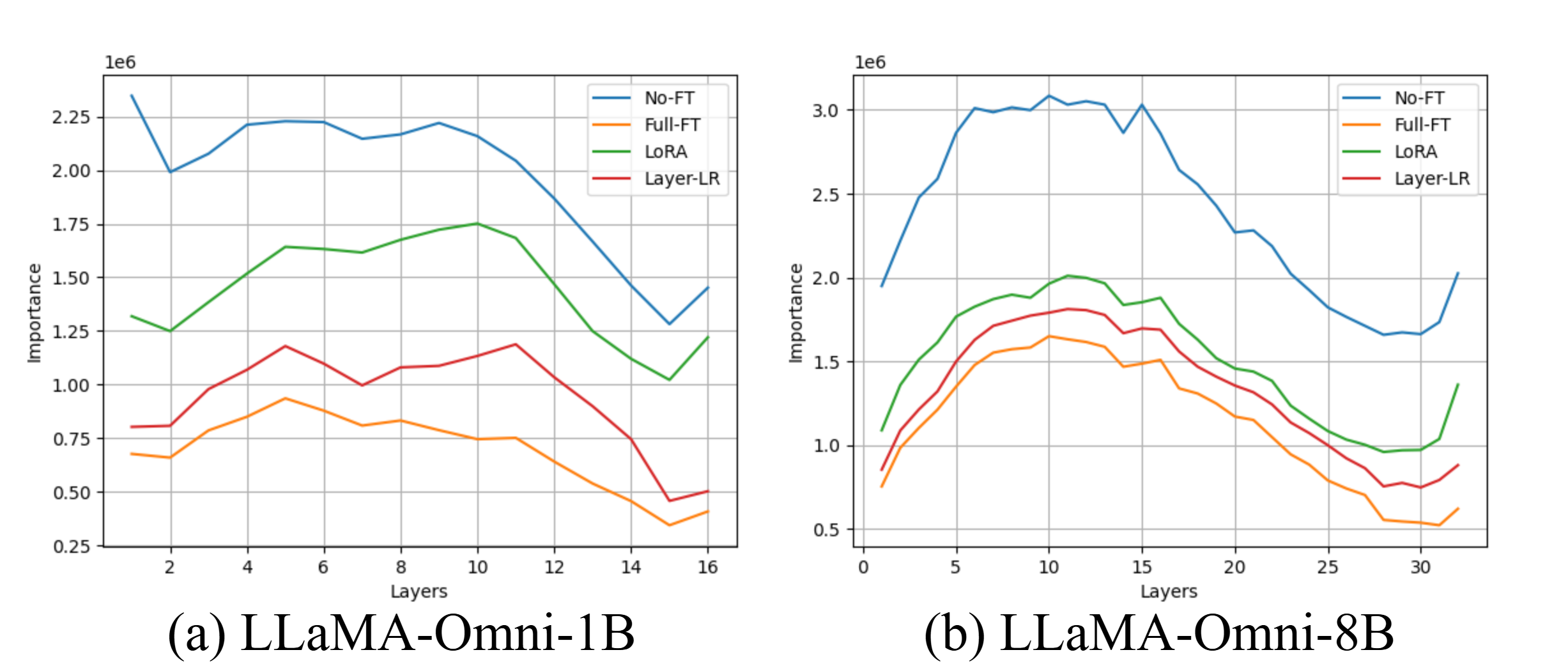}
    \caption{Layer-wise distribution of textual parameter importance before and after different fine-tuning processes.}
\label{fig:improved_layer_dist}
\end{figure}
% \begin{figure}[tb]
% %
% \begin{minipage}[t]{0.48\linewidth}
%   \centering
%   \centerline{\includegraphics[width=4.0cm]{figures/tensor_imp_layers_1b_2.png}}
% %  \vspace{1.5cm}
%   \centerline{(a) speech}\medskip
% \end{minipage}
% \begin{minipage}[t]{0.48\linewidth}
%   \centering
%   \centerline{\includegraphics[width=4.0cm]{figures/tensor_imp_layers_8b_2.png}}
% %  \vspace{1.5cm}
%   \centerline{(b) text}\medskip
% \end{minipage}
% %
% \caption{Layer-wise distribution of textual parameter importance before and after different fine-tuning processes.}
% \label{fig:improved_layer_dist}
% %
% \end{figure}

\subsection{Analysis}
\label{ssec:analysis}

\subsubsection{Mitigation of Distribution Shift}
% To further investigate the impact of different fine-tuning methods on the model's textual capabilities, we visualized the layer-wise distribution of textual parameter importance before and after applying each fine-tuning method. 
To examine why the proposed strategies are effective, we analyzed their impact on model's importance distribution. 
As shown in Figure \ref{fig:improved_layer_dist}, 
% in both the 1B and 8B models, the layer-wise distributions of textual parameter importance in models fine-tuned with the layer-wise learning rate method and LoRA more closely resemble that of the pre-fine-tuned model. This pattern indicates that both methods mitigate the issue observed in full fine-tuning, where the textual parameter importance declines more rapidly in deeper layers, thereby demonstrating their superior ability to preserve the model's original textual capabilities.
both layer-wise learning rate scheduling and LoRA produce distributions closely resembling those of the original pre-trained model, whereas full fine-tuning substantially distorts them. This confirms that both strategies mitigate the distribution shift identified in Section \ref{sssec:text_dis_shift}, thereby preserving the model’s textual competence.

% Additionally, we visualized the heatmaps of parameter changes within internal modules of both the Speech-Llama-8B-Layer-LR and Speech-Llama-8B-LoRA. As shown in Figure \ref{fig:param_change}, we observed that parameter changes in models fine-tuned with both methods exhibit row- and column-wise clustering patterns, with this effect being more pronounced in Speech-Llama-8B-LoRA. This observation aligns with the rank clustering phenomenon discussed in Section \ref{ssec:dim_dep} and supports our hypothesis presented in Section \ref{ssec:improve} that the optimization introduced by LoRA fine-tuning also demonstrates structured clustering characteristics—making it better suited to adapt to speech input requirements. This may explain why Speech-Llama-8B-LoRA achieves higher accuracy on speech question answering compared to Speech-Llama-8B-Layer-LR.
% Compared to full fine-tuning, LoRA improves the transfer efficiency from textual to speech capabilities. Although the layer-wise learning rate method preserves the model's original textual ability more effectively, its lower transfer efficiency ultimately results in inferior speech comprehension performance relative to LoRA fine-tuning.
\subsubsection{Comparative Analysis of Mitigation Strategies}
\begin{figure}[t]
    \centering
    \includegraphics[width=0.8\linewidth]{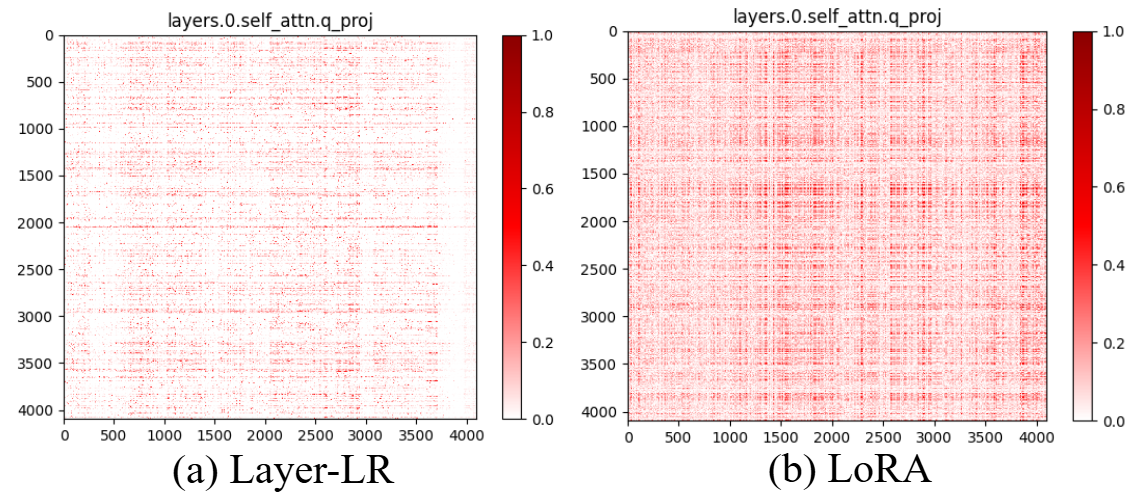}
    \caption{Heatmap of parameter changes (after normalization) under different fine-tuning methods.}
    \label{fig:param_change}
\end{figure}

We further compared the two strategies by visualizing the parameter changes induced by each method. As illustrated in Figure \ref{fig:param_change}, LoRA's updates exhibit a more pronounced row-and-column clustering pattern, consistent with the rank clustering phenomenon observed in Section \ref{ssec:dim_dep}. This supports that LoRA adapts more efficiently to the model’s inherent low-rank knowledge structure, which explains its superior performance on spoken QA. In contrast, layer-wise learning rate scheduling applies more conservative updates that effectively preserve textual knowledge but yield less efficient transfer to the speech modality, accounting for its comparatively weaker performance on S2T tasks.

\subsubsection{Ablation Study on LoRA Rank}
We also conducted an ablation study on the LoRA rank parameter $r$. As shown in Table \ref{tab:rank_ablation}, increasing $r$ from 8 to 16 consistently improves S2T performance while maintaining stable T2T results. However, a further increase to $r=24$ leads to a noticeable drop in both T2T and S2T accuracy, likely due to overfitting or excessive disruption of pre-trained knowledge. These results indicate that moderate ranks (e.g., $r=16$) strike a better balance between preserving textual competence and enabling effective speech adaptation.

% \begin{figure}[t]
% %
% \begin{minipage}[t]{0.48\linewidth}
%   \centering
%   \centerline{\includegraphics[width=4.0cm]{figures/params_change_module_dist_fullFT.png}}
% %  \vspace{1.5cm}
%   \centerline{(a) speech}\medskip
% \end{minipage}
% \begin{minipage}[t]{0.48\linewidth}
%   \centering
%   \centerline{\includegraphics[width=4.0cm]{figures/params_change_module_dist_LoRA.png}}
% %  \vspace{1.5cm}
%   \centerline{(b) text}\medskip
% \end{minipage}
% %
% \caption{Heatmap of parameter changes under different fine-tuning methods.}
% \label{fig:param_change}
% %
% \end{figure}

\begin{table}[t]
    \centering
    \caption{Results on the Llama Questions and Web Questions benchmarks under different LoRA rank settings.}
    \begin{tabular}{c c c c c}
        \hline
        \multirow{2}{*}{Rank} & \multicolumn{2}{c}{Llama Questions} & \multicolumn{2}{c}{Web Questions} \\
         & T2T & S2T & T2T & S2T \\ \hline 
         8 & 81.3 & 74.7 & 56.3 & 41.7 \\
         16 & 81.0 & 75.0 & 56.7 & 42.9 \\
         24 & 79.3 & 75.3 & 54.7 & 40.1 \\ \hline
    \end{tabular}
    \label{tab:rank_ablation}
\end{table}

% \begin{table}[t]
%     \centering
%     \begin{tabular}{c c c c c}
%         \hline
%         \multirow{2}{*}{$\lambda$} & \multicolumn{2}{c}{Llama Questions} & \multicolumn{2}{c}{Web Questions} \\
%          & T2T & S2T & T2T & S2T \\ \hline 
%          0.9 &  & 66.3 &  & 28.9 \\
%          0.6 &  & 66.3 &  & 29.2 \\
%          0.4 & 73.7 & 68.3 & 54.7 & 30.2 \\ \hline
%     \end{tabular}
%     \caption{Results on the question-answering benchmarks under different settings. }
%     \label{tab:lambda_ablation}
% \end{table}

\section{Conclusion}
\label{sec:conclusion}

% In this paper, we employed parameter importance method to analyze the internal parameters and fine-tuning processes of speech LLMs. Analytical results indicates that the reason for the degradation in textual capability of speech LLMs after fine-tuning is textual importance distribution shift. Based on the analysis, we propose two methods to mitigate the damage to textual ability during fine-tuning: layer-wise learning rate scheduling and LoRA. These methods effectively mitigate the degradation of textual capability during fine-tuning, enhance the text-based question answering performance of speech LLMs, and consequently lead to improvement in their speech-based question answering capability.
This paper investigated the degradation of textual capabilities in speech LLMs through a parameter-level analysis. We identified the primary mechanism as a textual importance distribution shift, where fine-tuning for speech disrupts the model's original knowledge structure. Based on this finding, we adopted two mitigation strategies: layer-wise learning rate scheduling and LoRA. Experimental results confirmed that both methods effectively preserve textual competence by alleviating this distribution shift, which in turn leads to notable improvements in spoken QA performance. For future work, we plan to extend our analytical framework to other speech LLM paradigms, such as those based on vocabulary expansion, to assess the generality of our findings. 
% Another promising direction is to develop more advanced strategies that better balance knowledge preservation with adaptation efficiency.

% \clearpage

%\vfill
%\pagebreak

\bibliographystyle{IEEEbib}
\bibliography{refs}

\end{document}